MDPI

*Article*

# Machine Learning Sensors for Diagnosis of COVID-19 Disease Using Routine Blood Values for Internet of Things Application

**Andrei Velichko** [1,*], **Mehmet Tahir Huyut** [2,*], **Maksim Belyaev** [1], **Yuriy Izotov** [1] and **Dmitry Korzun** [3]

[1] Institute of Physics and Technology, Petrozavodsk State University, 33 Lenin Ave.,
185910 Petrozavodsk, Russia
[2] Department of Biostatistics and Medical Informatics, Faculty of Medicine,
Erzincan Binali Yıldırım University, 24000 Erzincan, Türkiye
[3] Department of Computer Science, Institute of Mathematics and Information Technology, Petrozavodsk
State University, 33 Lenin Ave., 185910 Petrozavodsk, Russia
* Correspondence: velichko@petrsu.ru (A.V.); tahir.huyut@erzincan.edu.tr (M.T.H.)

**Abstract:** Healthcare digitalization requires effective applications of human sensors, when various parameters of the human body are instantly monitored in everyday life due to the Internet of Things (IoT). In particular, machine learning (ML) sensors for the prompt diagnosis of COVID-19 are an important option for IoT application in healthcare and ambient assisted living (AAL). Determining a COVID-19 infected status with various diagnostic tests and imaging results is costly and time-consuming. This study provides a fast, reliable and cost-effective alternative tool for the diagnosis of COVID-19 based on the routine blood values (RBVs) measured at admission. The dataset of the study consists of a total of 5296 patients with the same number of negative and positive COVID-19 test results and 51 routine blood values. In this study, 13 popular classifier machine learning models and the LogNNet neural network model were exanimated. The most successful classifier model in terms of time and accuracy in the detection of the disease was the histogram-based gradient boosting (HGB) (accuracy: 100%, time: 6.39 sec). The HGB classifier identified the 11 most important features (LDL, cholesterol, HDL-C, MCHC, triglyceride, amylase, UA, LDH, CK-MB, ALP and MCH) to detect the disease with 100% accuracy. In addition, the importance of single, double and triple combinations of these features in the diagnosis of the disease was discussed. We propose to use these 11 features and their binary combinations as important biomarkers for ML sensors in the diagnosis of the disease, supporting edge computing on Arduino and cloud IoT service.

## 1. Introduction

Identified in 2019, COVID-19 is an infectious disease caused by the novel severe acute respiratory syndrome coronavirus (SARS-CoV-2) [1,2]. Since the World Health Organization (WHO) declared the SARS-CoV-2 infection as a pandemic, the epidemic still maintains its severity to this day [3,4]. The early diagnosis of patients is extremely important to manage this unprecedented emergency [5,6]. The preferred gold standard method for detecting SARS-CoV-2 infections is the reverse polymerase chain reaction (PCR) or reverse transcriptase-PCR (RT-PCR) technique [7]. However, the execution of the test is time consuming (not less than 4–5 h under optimum conditions) and many favorable conditions must be met, such as the use of special equipment and reagents, the collection of samples and the necessity of trained personnel [8]. Machine learning (ML) and artificial intelligence (AI) models provide a powerful motivation to uncover insights from patients' data in tragic events such as the COVID-19 pandemic or in situations wherein guidelines have not yet been created [9]. ML and AI methods select the relevant biomarkers, revealing their predictive importance and consistently detecting their interactions with each other. Moreover, the diagnostic performance of these methods has the ability to be improved [9–11]. AI studies for the early detection, diagnosis and prognosis of COVID-19 relied on computed tomography (CT) and RBVs. However, imaging-based solutions are costly and require specialized equipment. Machine learning (ML) and AI studies based on RBVs features are a more economical and rapid alternative method for the early detection, diagnosis and prognosis of COVID-19 [7,11,12]. Previous studies have indicated that this disease can accompany multi-organ dysfunction and cause a variety of symptoms [3,13–15]. COVID-19 can cause severe pneumonia and severe ARDS due to inflammatory cytokine storms [5,14]. The excessive and uncontrolled release of proinflammatory cytokines was considered the most important primary cause of death, similar to other infections caused by pathogenic coronaviruses [16].

The pathogen may require special attention in intensive care units (ICUs) and cause a serious respiratory disorder, in some cases leading to death [14,16]. Moreover, it is difficult to distinguish symptoms of COVID-19 from known infections in the majority of patients [14,17,18]. This predictive analytics is especially required in medical information systems (MISs) to support clinical or managerial decisions.

COVID-19 may be part of a broader spectrum of hyperinflammatory syndromes characterized by the cytokine release syndrome (CRS), such as secondary hemophagocytic lymphohistiocytosis (sHLH) [19–21]. The activation of the monocyte–macrophage system just before the disease leads to pneumonia [22,23]. During this period, changes in many routine laboratory parameters such as D-dimer and fibrinogen have been reported in COVID-19 patients [1,2,4,5,14,22,24]. High ferritin, D-dimer, lactate dehydrogenase and IL-6 levels are indicators of poor prognosis and risk of death in patients [25–27]. In addition, Winata and Kurniawan [28] reported increased D-dimer and fibrinogen degradation product (FDP) in all patients in the late stage of COVID-19. This indicates that D-Dimer and FDP levels are elevated due to increased hypoxia in severe COVID-19 conditions and are significantly associated with coagulation. Kurniawan et al. [29] reported that hyperinflammation, coagulation cascade, multi-organ failure, which play a role in the etiopathogenesis of COVID-19, and biomarkers associated with these conditions, such as CRP, D-Dimer, LDH and albumin, may be useful in predicting the outcome of COVID-19.

The previous studies detected the clinical significance of changes in the routine blood values (RBVs) in the diagnosis and prognosis of infectious diseases [1,2,4,5,30,31]. However, Jiang et al. [32], Zheng et al. [33] and Huyut [11] noted that information on early predictive RBVs should be supplemented with large samples, especially for severe and fatal cases of COVID-19.

The uncontrolled spread of the disease in pandemics distresses health systems. The early detection of patients in pandemics is an important but clinically difficult process in terms of morbidity and mortality [14,24]. The diagnosis and prognosis of COVID-19 with the use of advanced devices can provide support in improving patient comfort, health system and tackling economic inadequacies [6,11,12]. In this context, studies are carried out to diagnose and determine the severity of the disease in the early period by using ML and AI-based methods as well as RBVs data [7,11,12]. The basic element in ML approaches is to determine the feature vector with a linear classifier [30]. Since ML algorithms require a sufficiently large number of samples, the problem of dimensionality in these methods is inevitable. To minimize this problem, the dataset should be reduced by finding a less dimensional attribute matrix. The dimensionality problem can be minimized by discarding irrelevant features with the feature selection procedure [30,31].

Feature selection methods can be summarized under three main headings: embedded methods, filters and wrappers (backward elimination, forward selection, recursive feature elimination) [30,31]. Feature selection in embedded methods is part of the training process and, therefore, this method lies between filters and wrappers. In the embedded methods, the determination of the best subset of features is performed during the training of the classifier (for example, when optimizing weights in a neural network). In terms of computational cost, embedded methods are more economical than wrappers [30].

Although we can find many case studies for all three feature selection methods, most feature selection methods are filters [30]. The existence of a large number of available feature selection methods complicates the selection of the best method for a particular problem [31]. The popular feature selection methods include correlation-based feature selection (CFS) [34], consistency-based filtering [35], INTERACT [36], information gain [37], ReliefF [38], recursive feature elimination for support vector machines (SVM-RFE) [39], Lasso editing [40] and minimum redundancy maximum relevance (mRMR) algorithm (developed specifically for dealing with microarray data) [30].

We examined the SARS-CoV-2-RBV1 database using the LogNNet neural network [12]. LogNNet can be defined as a feed forward network that increases the classification accuracy by chaotic mapping that fills a reservoir matrix. It is important to optimize the chaotic map parameters in data analysis by applying the LogNNet neural network. In addition, by taking advantage of chaotic mapping, it is possible to significantly reduce the RAM usage by a neural network. These results show that LogNNet can be used effectively in Internet of Things (IoT) mobile devices.

The main point for many digital health solutions during the pandemic process is the production of effective, fast and inexpensive alternative methods for the early diagnosis and treatment of COVID-19 patients. However, even the most knowledgeable and experienced physicians can interpret little of the information contained in routine blood laboratory results, and it is extremely difficult to determine the severity of COVID-19 patients based on RBVs findings alone [41]. In this context, ML classification models run with RBV-based data to determine the preliminary diagnosis of COVID-19 can be an effective tool in clinical decision support systems with an accuracy of over 95%. In this study, 13 ML models and LogNNet neural networks were applied in the diagnosis of suspected cases with an alternative device, based on LogNNet and Andrunio solutions, as only RBV-based, and the most important features were determined. We made a clinical interpretation of the relationship between these features and their various combinations with the disease. We achieved the performance of all models in detecting the diagnosis of the disease and reached up to 99.8% accuracy. ML sensors (Sensors 1.0 type) for the diagnosis of the COVID-19 disease have been successfully tested in the IoT environment, and the diagnosis of the disease has been implemented in offline and online modes. In offline mode, ML sensors were run on an Arduino board with a LogNNet neural network with a total RAM consumption of ~4 kB. Obtaining the findings in this study over a large sample is an important advantage in terms of the validity of the study. We believe that this study will help to identify suspected patients with a high probability of being infected with COVID-19 at the time of admission to the hospital with a fast and economical method, which will make important contributions to the detection of the disease before it progresses.

The paper has the following structure. Section 2 present the related studies, Section 3 describes the data collection procedure, correlation analysis of features, machine learning methods and the implementation of LogNNet on an Arduino board. Section 4 presents the results from the correlation analysis of dataset, classification results, one, double, triple and 11 feature combinations in the detection of sick and healthy individuals, and the ML sensor concept for IoT. Section 5 discusses the results and compares them with known developments. Section 6 presents the limitations of the study. In conclusion, a general description of the study and its scientific significance are given.

## 2. Related Studies

The prompt diagnosis of COVID-19 seems to be a promising advancement for applying at-home health care and AAL [42]. The digitalization of healthcare calls for effective applications of human body sensors [43] and human sensing [44], including ML sensors, to continuously monitor various parameters of the human body in everyday life with the help of the IoT [45]. Everyday human body sensors testify to the growing number of applications in IoT-enabled ambient intelligence (AmI) systems [46]. The paradigms of ML sensors [47] and artificial intelligence (AI) sensors [48,49] are similar in meaning. The ML sensor paradigm was further developed by Warden et al. [47] and Matthew Stewart [50], wherein the authors introduced the terms Sensors 1.0 and Sensors 2.0 devices. Sensors 2.0 devices are a combination of both a physical sensor and a machine learning module in one package. Sensors 2.0 devices process data internally, ensuring data security, while in Sensors 1.0 devices, these modules are physically separated. In addition, the authors proposed the concept of creating a dataseet of ML sensors. Therefore, the development of technology for creating ML sensors for the diagnosis of the COVID-19 disease is an urgent problem.

In previous studies, the RBVs of people who lived and died from COVID-19, or patients with COVID-19 and healthy individuals, were statistically compared [1,3,14,22–24,26,51]. In addition, differences in many RBVs characteristics are known between mild and severe COVID-19 patient groups according to statistical methods. However, this study demonstrates that ML models using only one or two features can detect COVID-19 patients from a large group of patients with high accuracy. Therefore, this study will be an alternative approach with extremely high sensitivity for the diagnosis of COVID-19. ML algorithms allow for an easy interpretation of complex association structures in data by simultaneously evaluating the cumulative effects of numerous biomarkers to discover higher-order interactions [4,6,9]. With this benefit, the strengths of using ML in clinical medicine are considered as an opportunity. Although various clinical studies [7,52,53] have highlighted how blood test-based diagnosis can provide an effective and low-cost alternative for the early detection of COVID-19 cases, relatively few ML models have been applied to blood parameters [7,54].

An evaluation of lung CT images to predict lung cancer using deep learning with an improved abundant clustering technique and instant trained neural networks approach was performed in [55]. The authors achieved an accuracy of up to 98.42% in cancer diagnosis with a minimum classification error of 0.038. Cui et al. [56] examined the distribution of pixels in the images with the fuzzy Markov random field segmentation approach using positron emission tomography (PET) and computed tomography (CT) images of the affected area associated with lung tumor. The developed method provided an accuracy of 0.85 in recognizing the lung tumor region. Tomita et al. [57] ran a logistic regression (LR) analysis , support vector machine (SVM) and deep neural network (DNN) models with biochemical findings, lung function tests and bronchial challenge test features to predict the initial diagnosis of adult asthma. In the pre-diagnosis of adult asthma, the DNN model showed 0.98%-ACC, the SVM model 0.82%-ACC and the LR model 0.94%-ACC. Ryu et al. [58] used various ML models and a deep neural network model for the pre-diagnosis of diabetes mellitus using various physical and routine blood values features. The deep neural network has been the most successful model with a value of 0.80-AUC in diabetes mellitus. Kolachalama et al. [59] used a six-convolutional deep learning architecture (CNN) with histological images, biopsy results and some clinical phenotypes to classify kidney disease severity. The CNN model was found to be more successful with AUC values of 0.878, 0.875 and 0.904, respectively, than the pathologist-predicted fibrosis score (0.811, 0.800 and 0.786 AUC, respectively) for assessing 1-, 3- and 5-year renal survival. In a study conducted to identify patients at risk of early diagnosis of fatty liver disease, Wu et al. [60] used an artificial neural network model with three ML models. The most successful model in the diagnosis of risky patients was the random forest with 87.48-ACC and 0.92-AUC values. Oguntimilehin et al. [61] used an ML technique on a set of labeled typhoid fever contingent variables for the diagnosis of typhoid fever and to establish explicable rules. The labeled database is divided into five different levels of typhoid fever severity, with classification accuracies on both the training set and the test set of 95% and 96%, respectively. The application was implemented using Visual Basic as the front-end and MySQL as the back-end. Kouchaki et al. [62] used various ML methods to predict the resistance to MTB in Mycobacterium tuberculosis (MTB) patients given a specific drug in a timely manner and to identify resistance markers. Compared to the traditional molecular diagnostic test, the AUC values of the best ML classifiers were higher for all drugs. Logistic regression and gradient tree reinforcement methods performed better than other techniques. Taylor et al. [63] ran six machine learning algorithms with 10 features consisting of patient demographics, RBVs results and drug information for the diagnosis of and treatment decisions for urinary tract infection. The best performing model, XGBoost, diagnosed the presence of a urinary tract infection with a high AUC value (0.826–0.904 confidence interval).

Yang et al. [64] ran four ML models on 3356 patients (42% COVID-19 positive) using 27 features covering both blood count and biochemical parameters. A gradient boosted decision tree model was the most successful model in the diagnosis of the disease with a value of 0.85-AUC. Booth et al. [9] operated 26 RBVs data elements with a support vector machine to detect COVID-19 patients at high mortality risk and determined prognostic biomarkers with a value of 0.93-AUC. Huyut [11] classified severely and mildly infected patients from a large population of COVID-19 patients using 12 supervised ML models and 28 routine blood values. The models with the highest AUC for identifying mildly infected patients were found to be local weighted learning at 0.95%, Kstar at 0.91%, Naive bayes at 0.85% and K nearest neighbor at 0.75%. Brinati et al. [65] ran 13 RBVs with various ML methods to detect COVID-19 patients (102 negative and 177 COVID-19 positive). They noted that the models with the highest accuracy in the diagnosis of the disease were random forest (82%) and logistic regression (78%). Huyut and Velichko [12] determined the diagnosis and prognosis of the COVID-19 disease by running the LogNNet neural network model on 51 RBVs features. The model achieved an accuracy of 99% in the diagnosis of the disease and 84% in its prognosis. Zhang et al. [66] used a variety of demographic indicators and 21 RBVs using a Lasso-based neural network model to detect predictors of mortality from COVID-19. The success of the model in determining the clinical status of the patients was 98%-AUC. Alle et al. [67] applied the XGboost and logistic regression model on a dataset of various clinical and laboratory tests to predict COVID-19 mortality and found accuracy rates of 83% and 92%, respectively. Gao et al. [68] applied an ensemble model derived from support vector machine (SVM), gradient augmented decision tree (GBDT) and neural network (NN) algorithms using 28 immune/inflammatory features to detect COVID-19. The developed model reached 0.99 AUC in detecting infected patients. Vaishnav et al. [69] used various machine learning models to predict mortality from COVID-19, and the decision tree regression model produced a 70% accuracy and the random forest regression model a 76% accuracy. Huyut and İlkbahar [5] used various biomarkers with the CHAID decision tree to detect the diagnosis and prognosis of COVID-19. The model produced an 81.6% accuracy in recognizing the disease and a 93.5% accuracy in determining the prognosis of the disease. Huyut and Üstündağ [6] used 23 blood gas parameters with the CHAID decision tree to predict the diagnosis and prognosis of COVID-19. The model produced a 68.2% accuracy in recognizing the disease and a 65.0% accuracy in determining the prognosis of the disease. Kukar et al. [70] constructed a machine learning model based on 35 RBVs to diagnose 5333 negative and 160 positive COVID-19 patients with various bacterial and viral infections. The model showed an 81.9% sensitivity and a 97.9% specificity in detecting patients. Mei et al. [71] developed a model combining CNN and multilayer sensor to detect COVID-19 using computed tomography (CT), various clinical information elements and some RBVs data. The model reached an 84% sensitivity and an 83% specificity in recognizing the disease.

AI studies on the risk of poor outcome for COVID-19 patients need further validation with larger samples [11,25,72]. Furthermore, previous AI studies using RBVs for COVID-19 diagnosis and prognosis which covered the early stages of the outbreak included less blood values and reported poorer performance. Therefore, to detect the disease in the later stages of the epidemic, it is necessary to study ML models on a larger sample, which can achieve higher accuracy and use most RBVs.

## 3. Data and Methods

The data used in this study were collected retrospectively from the information system of Erzincan Binali Yıldırım University Mengücek Gazi Training and Research Hospital (EBYU-MG) between April and December 2021. The data used in this study are shared as open access under the name of "SARS-CoV-2-RBV1" in [12].

During the dates covered by this study, a diagnosis of SARS-CoV-2 was made by real-time reverse transcriptase polymerase chain reaction (RT-PCR) on nasopharyngeal or oropharyngeal swabs at the EBYU-MG hospital. RBVs results at first admission were recorded to prevent various complications.

### 3.1. Characteristic of Participants, Workflow and Datasets

Between the specified dates, the digital system of our hospital was scanned and patients diagnosed with COVID-19 (n = 2648) were selected from a large patient population (a dataset of approximately 80 thousand patients was scanned). The routine laboratory information of these patients was examined. The parameters (features) that were measured from at least 80% of the patients were used. Missing data were completed with the mean of the distribution and normalized. A total of 51 routine blood values calibrated from approximately 70 parameters were recorded. In addition, a group (control group) with the same number of negative COVID-19 tests (n = 2648) was identified and 51 characteristics of these individuals were recorded. Our control group arrived at the hospital only with the suspicion of COVID-19. Chronic disease information of the patients could not be reached. Only data of individuals over the age of 18 were recorded.

These two datasets were combined and named "SARS-CoV-2-RBV1" dataset. The SARS-CoV-2-RBV1 dataset includes immunological, hematological and biochemical RBVs parameters and consists of 51 features (Table 1). In the SARS-CoV-2-RBV1 dataset, positive COVID-19 test results were coded as 1 and negative test results as 0 (COVID-19 = 1, non-COVID-19 = 0).

The features in this dataset are calibrated and contain almost all of the routine blood values that are the subject of studies on COVID-19 in the literature. Therefore, we believe that the bias of our study using this dataset was minimized in comparison to the literature. In addition, the use of our dataset, which we share as open access, is important in terms of showing the reproducibility and auditability of the results.

**Table 1.** Feature numbering for SARS-CoV-2-RBV1 dataset [12].

| № | Feature | № | Feature | № | Feature | № | Feature | № | Feature |
|---|---------|---|---------|---|---------|---|---------|---|---------|
| 1 | CRP | 12 | NEU | 23 | MPV | 34 | GGT | 45 | Sodium |
| 2 | D-Dimer | 13 | PLT | 24 | PDW | 35 | Glucose | 46 | T-Bil |
| 3 | Ferritin | 14 | WBC | 25 | RBC | 36 | HDL-C | 47 | TP |
| 4 | Fibrinogen | 15 | BASO | 26 | RDW | 37 | Calcium | 48 | Triglyceride |
| 5 | INR | 16 | EOS | 27 | ALT | 38 | Chlorine | 49 | eGFR |
| 6 | PT | 17 | HCT | 28 | AST | 39 | Cholesterol | 50 | Urea |
| 7 | PCT | 18 | HGB | 29 | Albumin | 40 | Creatinine | 51 | UA |
| 8 | ESR | 19 | MCH | 30 | ALP | 41 | CK | | |
| 9 | Troponin | 20 | MCHC | 31 | Amylase | 42 | LDH | | |
| 10 | aPTT | 21 | MCV | 32 | CK-MB | 43 | LDL | | |
| 11 | LYM | 22 | MONO | 33 | D-Bil | 44 | Potassium | | |

CRP: C-reactive protein; INR: international normalized ratio; PT: prothrombin time; PCT: procalcitonin; ESR: erythrocyte sedimentation rate; aPTT: activated partial prothrombin time; LYM: lymphocyte count; NEU: neutrophil count; PLT: platelet count; WBC: white blood cell count; BASO: basophil count; EOS: eosinophil count; HCT: hematocrit; HGB: hemoglobin; MCH: mean corpuscular hemoglobin; MCHC: mean corpuscular hemoglobin concentration; MCV: mean corpuscular volume; MONO: monocyte count; MPV: mean platelet volume; PDW: platelet distribution width; RBC: red blood cells; RDW: red cell distribution width; ALT: alanine aminotransaminase; AST: aspartate aminotransferase; ALP: alkaline phosphatase; CK-MB: creatine kinase myocardial band; D-Bil: direct bilirubin; GGT: gamma-glutamyl transferase; HDL-C: high-density lipoprotein cholesterol; CK: creatine kinase; LDH: lactate dehydrogenase; LDL: low-density lipoprotein; T-Bil: total bilirubin; TP: total protein; eGFR: estimating glomerular filtration rate; UA: uric acid.

### 3.2. Correlation Analysis of Features

To determine the level of correlation between diagnosis and biochemical blood parameters, the original dataset was analyzed using the point-biserial correlation test [73]. Pearson correlation coefficient was calculated for each feature–feature pair, and a correlation matrix was compiled. The correlation matrix makes it possible to judge the strength and structure (positive or negative) of the linear relationship between diagnosis–feature and feature–feature pairs. The correlation matrix was created using the pandas software package [74].

### 3.3. Machine Learning Methods, Hyperparameters, Accuracy Estimation

Machine learning algorithms can be applied to a wide range of problems such as classification, clustering, regression analysis, time series forecasting, etc. [75]. The SARS-CoV-2-RBV1 dataset under study has an output parameter divided into two classes (positive or negative diagnosis for COVID-19), so the task of the machine learning algorithm is reduced to binary classification based on 51 features. This study compared the accuracy of the most popular binary classification algorithms: multinomial naive Bayes (MNB), Gaussian naive Bayes (GNB), Bernoulli naive Bayes (BNB), linear discriminant analysis (LDA), K-nearest neighbors (KNN), support vector machine classifier with linear kernel (LSVM), support vector machine classifier with non-linear kernel (NLSVM), passive-aggressive (PA), multilayer perceptron (MLP), decision tree (DT), extra trees (ETs) classifier, random forest (RF), histogram-based gradient boosting (HGB).

Each classifier model has hyperparameters, for which optimization is necessary to obtain the most accurate models. For optimization, the software package "auto-sklearn" [76] was used.

Before training the models, the initial data were subjected to preprocessing, which makes it possible to speed up the training of the models and improve the accuracy of the classification. Preprocessing includes two stages: (1) normalization of numerical values of the input data, (2) generation of additional features. Normalization is a procedure consisting of bringing numerical data to a single format, which has the following options: quantile transformer (QT)—transforms feature values so that they correspond to a uniform or normal distribution; robust scaler (RS)—subtracts the median values for each feature and scales according to the interquartile range; MinMax (MM)—scales feature values so that they are all in the range from the minimum to the maximum value. The procedure for generating additional features transforms the original set of features into a set of features with a different dimension. This helps to select the most important features, compose additional features from them or present the input data in a special format for the ML algorithm. The following methods for generating additional features were used: polynomial (PN)—creates features that are polynomial combinations of the original features; random trees embedding (RTE)—creates a multidimensional sparse feature representation, in which the data in each new feature are represented by binary values; extra trees preprocessor (ETP)—selects a part of the most important features that are evaluated using the extra trees algorithm; linear SVM preprocessor (LSVMP)—selects some of the most important features that are evaluated using the support vector machine algorithm; independent component analysis (ICA)—selects a set of statistically independent features from the entire original set; Nystroem sampler (NS)—transforms a set of initial features using a low-rank matrix approximation by the Nystroem method.

The accuracy of models $A_{NF}$ was assessed by the K-fold cross-validation method ($K = 5$) encapsulated in software packages, wherein the designation $A_{NF}$ refers to the classification accuracy when using $NF$ features. K-fold cross-validation method splits the original dataset into *K* parts and sequentially trains the model. One of the *K* parts of the dataset is used as a test sample, and the other parts as a training sample. Then, the obtained values of the classification accuracy on the test samples are averaged. The division of the base into parts is performed using stratification. Such approach makes it possible to reliably estimate the accuracy of models.

In this study, we used a less common ML algorithm based on the LogNNet neural network. The LogNNet 51:50:20:2 configuration was used, and a detailed configuration description is given in [75]. The LogNNet architecture is IoT-oriented and can run on devices with low computing resources (Section 3.4).

Each algorithm was given the same amount of time (1 h) to optimize the hyperparameters. A computer with an AMD Ryzen 9 3950X processor and 64 GB DDR-4 RAM was used to train the models.

### 3.4. Implementing LogNNet on an Arduino board

The Arduino Nano 33 IoT board was chosen as a prototype IoT edge device with limited computing resources. It is based on a 32-bit Microchip ATSAMD21G18 microcontroller with an ARM Cortex-M0+ computing core, a clock frequency of 48 MHz, 256 KB of flash memory and 32 KB of RAM. The neural network LogNNet 51:50:20:2 from [12] was programmed on the Arduino board and tested. Arduino Nano 33 IoT test circuit, LogNNet architecture and board are presented in Figure 1.

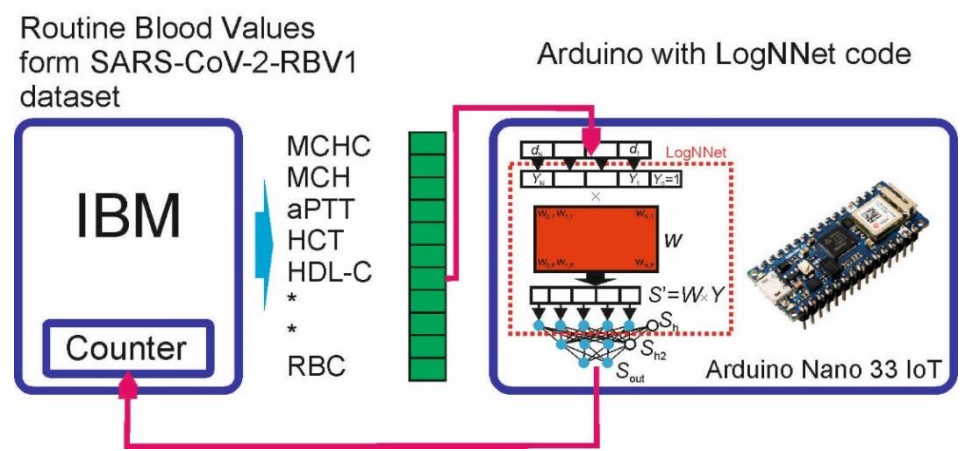

**Figure 1.** Arduino Nano 33 IoT test circuit and LogNNet architecture.

#### 3.4.1. LogNNet Program for Arduino Board

LogNNet transforms the input feature vector *d* into a normalized vector *Y*, which is multiplied with the reservoir matrix *W* filled with a chaotic mapping. We used the mapping congruent generator (1) with the parameters indicated in Table 2 and the data [12]. Then, the transformed vector passes the output classifier (two-layer feed-forward neural network with two hidden layers).

Let us denote the matrix of weight coefficients between the layers $S_h$ and $S_{h2}$ as $W_1$, and the matrix between the layers $S_{h2}$ and $S_{out}$ as $W_2$. At the output, there are two neurons for two classes (COVID-19 and non-COVID-19). Matrices of weight coefficients and values of normalization coefficients were calculated on a computer with high performance and saved in a separate library file. In addition, the library file (Supplementary Materials) contains the values of *K*, *D*, *L* and *C* required for calculating the *W* matrix, as well as data on configuration of the LogNNet 51:50:20:2 neural network.

**Table 2.** Chaotic map equation and list of parameters with limits.

| Chaotic Map | List of Parameters | Equation |
|---|---|---|
| Congruent generator | $K = 93$<br>$D = 68$<br>$L = 9276$<br>$C = 73$ | $\begin{cases} x_{n+1} = (D - K \cdot x_n) \mod L \\ x_1 = C \end{cases}$    (1) |

When LogNNet is running, the values of the elements of the matrix W (2550 values) are sequentially calculated using the congruent generator method (1) each time a feature vector is input. This approach does not store the matrix W in the RAM memory of the controller, and it leads to memory saving; however, it slows down the calculations of the neural network.

The Arduino IDE development environment was used to implement the algorithm. The library file with the matrices $W_1$ and $W_2$ and other coefficients necessary for the operation of the neural network were loaded at the beginning of the program. The complete code of the program is presented in Appendix A, Figure A1. The algorithm is divided into functions and procedures:

- Function "Fun_activ"—activation function, lines 10–12;
- Procedure "Reservoir"—calculation of coefficients of reservoir matrix W by congruent generator formula, multiplication of arrays and calculation of neurons in layer Sh, lines 14–28;
- Procedure "Hidden_Layer"—calculation of neurons in the hidden layer Sh2, lines 30–39;
- Function "Output_Layer_Layer"—calculation of the output layer Sout, lines 41–54;
- The "void loop" block is an executable loop, lines 61–77;
- "void setup" block—initialization block, lines 61–77.

The scaling factor "scale_factor = 1000" makes it possible to convert data from a floating point type to an integer (and vice versa), by multiplying (dividing) by a factor and rounding. In the Arduino, a float variable takes 4 bytes of RAM, and an integer variable takes 2 bytes of RAM. Therefore, storing matrices $W_1$, $W_2$ and other data in integer format are more efficient, and during library initialization, the data takes 2 times less RAM memory.

### 3.4.2. Test Scheme

Neural network testing is the serial sending of SARS-CoV-2-RBV1 data to the Arduino board and counting the correct network responses. The data is generated on an external computer (Figure 1). For sending data, a protocol was implemented that separates the elements of the feature vector $Y$ using the symbol "T" to avoid data gluing. At the end of the vector $Y$, special characters "FN" are placed, indicating the end of the data transfer. On the Arduino side, a protocol is implemented that recognizes the input data. In the "void loop" block, a loop is organized to check the availability of data in the serial port buffer using the Serial.Available function. This function returns "True" as soon as the Arduino receives data.

## 4. Results

### 4.1. Correlation Analysis of Dataset SARS-CoV-2-RBV1

Figure 2 presents the results of the correlation analysis of the diagnosis–feature and feature–feature pairs in the form of a "heatmap" over the entire volume of the SARS-CoV-2-RBV1 database. The first column of Table 3 shows the features most highly associated with the survivors and non-survivors of COVID-19 with a point-biserial correlation ($r_{pb}$) coefficient exceeding 0.5. Here, the negative or positive result of the point-biserial correlation coefficient provides information about the direction of the relationship between the diagnosis of the disease and the quantitative characteristics. As seen in the first column of Table 3, the features most associated with disease diagnosis are MCHC, HDL-C, cholesterol and LDH. The second column of Table 3 shows the accuracy of the cut-off values calculated by the threshold approach [12] for each trait in classifying COVID-19 patients. The features presented in the second column are the predictors that classify patients with the highest accuracy. For comparison, the results of the threshold classification $A_{th}$ from [12] are presented, and features with $A_{th} \geq 70\%$ are shown. The threshold classification method and the point-biserial correlation method give an intersecting set of features, but the threshold classification provides more diagnosis-related features. While the point-biserial correlation coefficient reveals the level of association between living and deceased COVID-19 patient traits, the diagnosis has only two values (1 and 0). However, when separating these two classes, the threshold method considers all the data of the relevant feature, and it has high sensitivity.

**Table 3.** Features most strongly correlated with the diagnosis according to the point-biserial correlation coefficient and the threshold correlation method.

| Feature (Point-Biserial Correlation Coefficient ($r_{pb}$)) | Feature (Threshold Accuracy of Classification $A_{th}$ from [12]) |
|---|---|
| MCHC (0.8) | MCHC (94.35%) |
| HDL-C (−0.77) | HDL-C (94.73%) |
| Cholesterol (−0.71) | Cholesterol (94.47%) |
| LDL (−0.68) | LDL (96.47%) |
|  | Triglyceride (90.96%) |
|  | Amylase (85.1%) |
|  | UA (81.12%) |
|  | TP (79.68%) |
|  | CK-MB (78.91%) |
|  | LDH (74.98%) |
|  | Albumin (72.91%) |

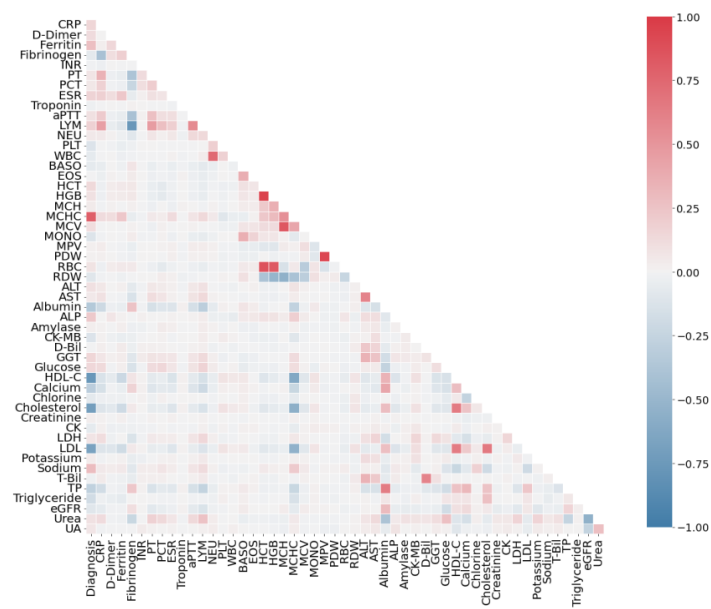

**Figure 2.** Correlation of the SARS-CoV-2-RBV1 dataset for diagnosis–feature (point-biserial correlation) and feature–feature pairs (Pearson coefficient).

An analysis of the correlation of features among themselves (Figure 2) reveals several features that are linearly dependent on each other. The most strongly correlated pairs with the Pearson coefficient exceeding 0.6 modulo are shown in Table 4. The same table presents Pearson's coefficients separately for a variety of COVID-19 positive and negative participants. Full heatmaps by class (COVID-19, non-COVID-19) are shown in Figures 3,4.

**Table 4.** The features most strongly correlated with each other by the Pearson coefficient for the entire database and separately for classes (positive or negative COVID-19 status).

| Pair Feature–Feature | Pearson's Coefficient for COVID-19 Diagnosis | Pearson's Coefficient for Positive COVID-19 | Pearson's Coefficient for Negative COVID-19 |
|---|---|---|---|
| | | Type High–High | |
| HCT–HGB | 0.96 | 0.95 (High) | 0.97(High) |
| MPV–PDW | 0.93 | 0.94 | 0.92 |
| HCT–RBC | 0.87 | 0.88 | 0.87 |
| MCH–MCV | 0.84 | 0.84 | 0.84 |
| HGB–RBC | 0.83 | 0.83 | 0.83 |
| NEU–WBC | 0.74 | 0.71 | 0.81 |
| Albumin–TP | 0.64 | 0.67 | 0.5 |
| MCH–MCHC | 0.53 | 0.62 | 0.99 |
| MCH–RDW | −0.55 | −0.61 | −0.51 |
| | | Type High–Low | |
| Fibrinogen–LYM | −0.77 | −0.78 (High) | −0.01 (Low) |
| Cholesterol–LDL | 0.65 | 0.59 | 0.012 |
| Cholesterol–HDL-C | 0.64 | 0.39 | −0.024 |
| Chlorine–Sodium | 0.18 | 0.63 | −0.025 |
| | | Type Low–High | |
| MCHC–MCV | 0.41 | 0.09 1(Low) | 0.84 (High) |
| ALT–AST | 0.6 | 0.48 | 0.76 |
| eGFR–Urea | −0.55 | −0.49 | −0.63 |
| INR–PT | 0.12 | 0.075 | 1 |
| D-Bil–T-Bil | 0.6 | 0.33 | 0.91 |
| HDL-C–LDL | 0.63 | 0.19 | 0.3 |

Three main types of pair correlations can be distinguished. The High–High type is the pairs of features for which the correlation has a high value does not depend on the presence or absence of COVID-19 disease. The High–Low type is the pairs of features that are highly correlated only in sick patients. The Low–High type is the pairs of features that are highly correlated only in healthy patients. In general, the features are more correlated in patients with COVID-19 (Figure 3). From a medical point of view, pair correlation will be reviewed in the Discussion Section.

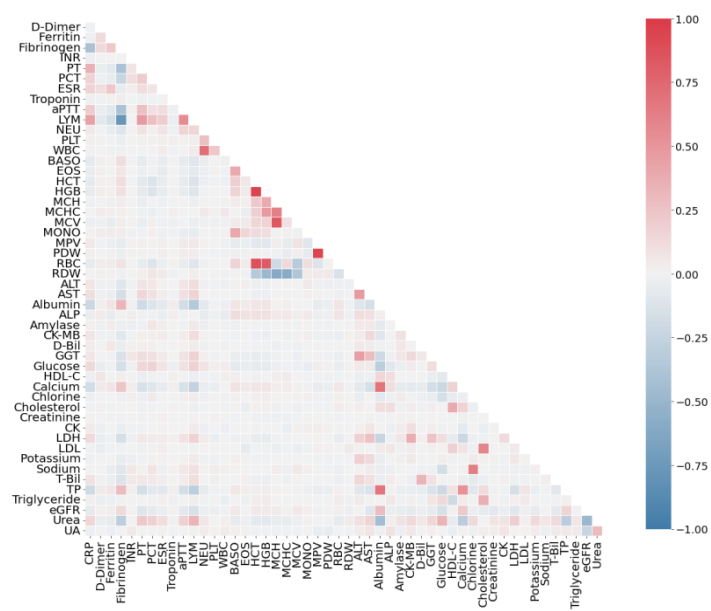

**Figure 3.** Pearson correlation analysis results for positive diagnoses for COVID-19 from the SARS-CoV-2-RBV1 dataset.

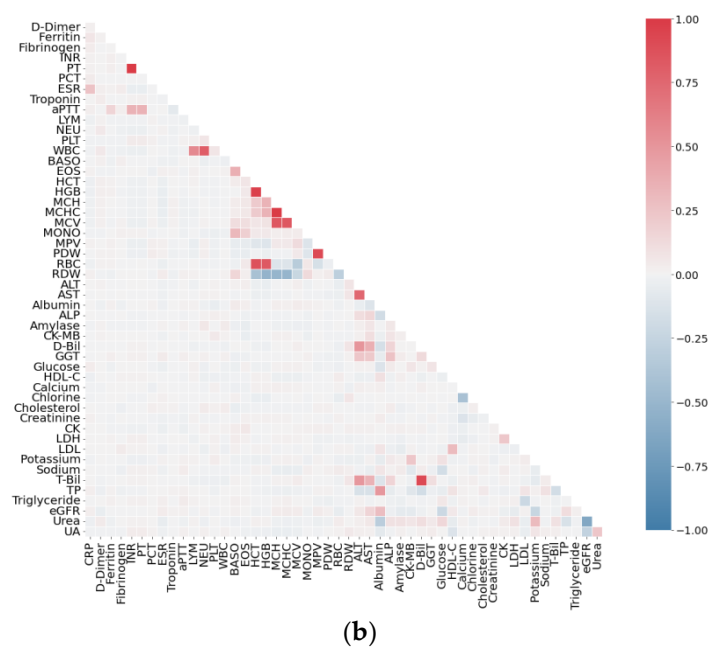

(**b**)

**Figure 4.** Pearson correlation analysis results for negative diagnoses for COVID-19 from the SARS-CoV-2-RBV1 dataset.

### 4.2. Classification Results for Dataset SARS-CoV-2-RBV1

Table 5 presents the results of the machine learning algorithms optimized to obtain the maximum classification values using 51 features. The results are sorted in descending order of algorithm efficiency. For each algorithm, the average training and inference time of the model and methods for preprocessing of the input data are given.

**Table 5.** Results of assessing the classification accuracy of machine learning models for the SARS-CoV-2-RBV1 dataset.

| Classification Algorithm | Average Model Accuracy $A_{51}$,% | Average Learning Time, s | Average Inference Time, μs | Normalization Method | Methods for Generating Additional Features |
|---|---|---|---|---|---|
| Histogram-based Gradient Boosting | 100 | 6.39 | 11.6 | - | - |
| Random Forest | 99.943 | 13.15 | 21.9 | QT | - |
| K-nearest neighbors | 99.924 | 3.17 | 22.1 | QT | ETP |
| Extra Trees classifier | 99.905 | 18.73 | 24.5 | RS | - |
| Multilayer Perceptron | 99.886 | 3.99 | 2.2 | RS | LSVMP |
| Multinomial Naive Bayes | 99.792 | 2.48 | 11.7 | QT | RTE |
| Linear Discriminant Analysis | 99.773 | 9.15 | 7.5 | QT | PN |
| Support Vector Machine with non-linear kernel | 99.754 | 222.41 | 43.2 | QT | NS |
| Decision Tree | 99.660 | 1.46 | 1.2 | RS | LSVMP |
| Passive-Aggressive | 99.641 | 2.91 | 13.1 | QT | RTE |
| Bernoulli Naive Bayes | 99.622 | 2.59 | 11.3 | QT | RTE |
| Support Vector Machine with linear kernel | 99.584 | 5.21 | 1242 | MM | PN |
| Gaussian Naive Bayes | 98.565 | 1.68 | 3.5 | QT | ICA |
| LogNNet [12] | 99.509 | 100 | 3 | - | - |

The accuracy of the algorithms $A_{51}$ ranged from 98.56% to 100%, indicating that all models were good at identifying the association of features with the diagnosis of COVID-19. The most efficient model is based on the histogram-based gradient boosting classifier with a 100% accuracy.

Figure 5 presents the learning curves for the HGB model using all the features from the dataset. The red curve (training accuracy) shows the training ability of the model, and the green curve (cross-validation accuracy) shows the generalization ability of the model depending on the number of training examples. Each point on the graph was obtained using five different splits into a test (20%) and training (80%) subsets.

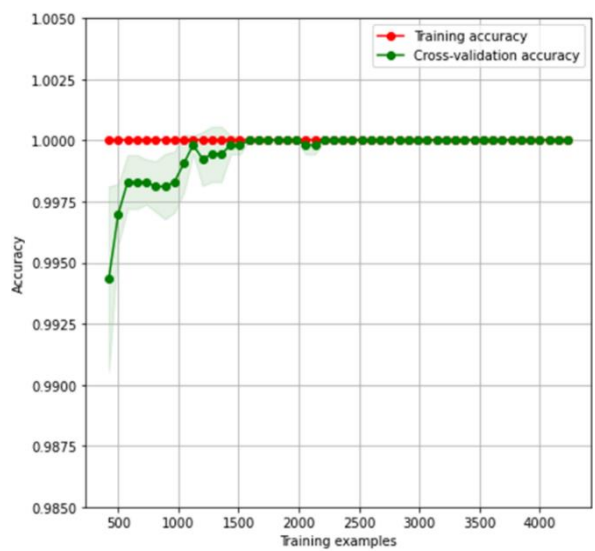

**Figure 5.** Learning curves for the histogram-based gradient boosting classifier model using 51 features from the SARS-CoV-2-RBV1 dataset.

The red curve represents the accuracy of the model on the training samples. The model has sufficient complexity to recognize all training samples with a 100% accuracy. The green curve represents the accuracy of the model on the test subset, the samples of which were not involved in model training. With an increase in training samples, the cross-validation accuracy of the model grows. The curves converge with each other and completely coincide when the number of training samples is more than 2500. The dots on the graph represent the average accuracy using five different splits, and the shaded areas represent the standard deviation.

Unlike other models, HGB does not require data preprocessing. The training time of the HGB model is about 6 s, which makes it possible to effectively use it to enumerate input features when searching for optimal combinations. The LogNNet model was used to implement the classification on the Arduino board, so its algorithm has a compact form suitable for IoT devices.

The HGB model was used to study the most significant combinations of the first, second and third features.

### 4.2.1. Investigation of the Effectiveness of the HGB Model Operating on one Feature

Table 6 presents the classification result of the SARS-CoV-2-RBV1 dataset for the HGB model using a single input feature. The features are sorted in descending order of $A_1$ classification accuracy. The most effective features are the first six features: LDL (№43), cholesterol (№39), HDL-C (№36), MCHC (№20), triglyceride (№48) and amylase (№31). The same features are dominant in assessing the correlation between the sign and the diagnosis from Table 3.

**Table 6.** Classification efficiency of SARS-CoV-2-RBV1 datasets using the single feature for the Histogram-based Gradient Boosting classifier.

| № | Feature | $A_1$,% | № | Feature | $A_1$,% | № | Feature | $A_1$,% | № | Feature | $A_1$,% |
|---|---------|---------|---|---------|---------|---|---------|---------|---|---------|---------|
| 43 | LDL | 96.84 | 4 | Fibrinogen | 76.03 | 50 | Urea | 68.10 | 21 | MCV | 56.43 |
| 39 | Cholesterol | 95.07 | 29 | Albumin | 75.3 | 7 | PCT | 63.25 | 22 | MONO | 56.26 |
| 36 | HDL-C | 94.99 | 44 | Potassium | 75.22 | 27 | ALT | 62.33 | 5 | INR | 56.19 |
| 20 | MCHC | 94.35 | 3 | Ferritin | 74.45 | 35 | Glucose | 62.17 | 6 | PT | 56.04 |
| 48 | Triglyceride | 93.76 | 38 | Chlorine | 73.18 | 49 | eGFR | 62.04 | 17 | HCT | 55.75 |
| 31 | Amylase | 90.01 | 46 | T-Bil | 72.77 | 14 | WBC | 61.91 | 26 | RDW | 55.62 |
| 51 | UA | 87.91 | 34 | GGT | 72.62 | 16 | EOS | 61.40 | 9 | Troponin | 54.07 |
| 42 | LDH | 85.76 | 41 | CK | 70.97 | 13 | PLT | 61.25 | 18 | HGB | 53.94 |
| 47 | TP | 80.41 | 2 | D-Dimer | 70.46 | 28 | AST | 60.55 | 25 | RBC | 53.43 |
| 37 | Calcium | 80.40 | 33 | D-Bil | 70.37 | 8 | ESR | 59.12 | 23 | MPV | 53.13 |
| 32 | CK-MB | 79.73 | 11 | LYM | 69.90 | 15 | BASO | 58.72 | 24 | PDW | 53.09 |
| 1 | CRP | 77.81 | 45 | Sodium | 69.35 | 12 | NEU | 57.51 | 19 | MCH | 52.13 |
| 30 | ALP | 77,71 | 40 | Creatinine | 69,24 | 10 | aPTT | 56,53 | | | |

### 4.2.2. Investigation of the Effectiveness of the HGB Model Operating on two Features

Table 7 presents the classification result of the SARS-CoV-2-RBV1 dataset for the HGB model using two input features. The pairs of features are sorted in descending order of classification accuracy $A_2$.

**Table 7.** Classification efficiency of SARS-CoV-2-RBV1 dataset using 2 features for the Histogram-based Gradient Boosting classifier.

| № | First Feature | Second Feature | Average Accuracy $A_2$,% |
|---|---|---|---|
| 20-19 | MCHC | MCH | 99.81 |
| 43-32 | LDL | CK-MB | 99.62 |
| 36-32 | HDL-C | CK-MB | 99.49 |
| 48-32 | Triglyceride | CK-MB | 99.45 |
| 43-39 | LDL | Cholesterol | 99.43 |
| 43-20 | LDL | MCHC | 99.22 |
| 39-36 | Cholesterol | HDL-C | 99.18 |
| 39-48 | Cholesterol | Triglyceride | 99.11 |
| 43-42 | LDL | LDH | 99.05 |
| 43-31 | LDL | Amylase | 99.03 |
| 36-20 | HDL-C | MCHC | 98.98 |
| 43-51 | LDL | UA | 98.86 |
| 36-31 | HDL-C | Amylase | 98.81 |
| 39-20 | Cholesterol | MCHC | 98.73 |
| 20-48 | MCHC | Triglyceride | 98.65 |
| 39-38 | Cholesterol | Chlorine | 98.62 |
| 43-38 | LDL | Chlorine | 98.43 |
| 20-31 | MCHC | Amylase | 98.28 |
| 36-42 | HDL-C | LDH | 98.16 |
| 48-42 | Triglyceride | LDH | 98.14 |

The resulting pairs contain the most effective features: LDL (№43), cholesterol (№39), HDL-C (№36), MCHC (№20), triglyceride (№48) and amylase (№31), which have the best $A_1$ score (Table 7). The best result ($A_2$ = 99.81) was obtained for the MCHC–MCH feature pair. At the same time, the pair contains the MCH (№19) feature with low efficiency ($A_1$ = 52.13) and Pearson correlation ~0.041. Such a combination of features with high and low correlation is observed very often, and this combination results in a high classification efficiency. Among the features from Table 7, the following have a low linear correlation with the diagnosis: MCH (0.041), UA (0.066), amylase (0.03) and LDH (0.071). Pearson's coefficient from the distribution in Figure 3 is indicated in brackets.

There are pairs consisting entirely of effective features, for example, LDL–MCHC (№43-№20), HDL-C–MCHC(№36-№20), etc. Figure 6 shows the relationship between the feature pairs for the top 50 results. The main six features are in the center. Asterisks indicate the features that most often form a pair with the main features: UA (№51), LDH (№42), CK-MB (№32) and ALP (№30). The main feature LDL (№43) forms the largest number of effective pairs for classification.

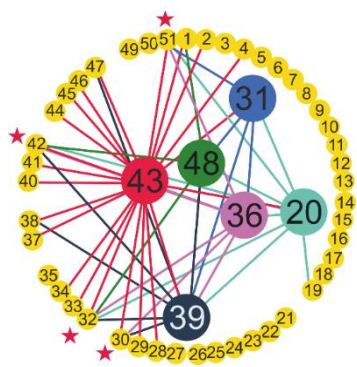

**Figure 6.** Pairs of features with high classification efficiency SARS-CoV-2-RBV1 dataset for the Histogram-based Gradient Boosting classifier.

To find the reasons for the effectiveness of the pairs of features from Table 7, two-dimensional distributions of the diagnosis (attractors) were constructed for the first six pairs (Figure 7). For the healthy patients (non-COVID-19), there are clear linear and cruciform attractors, while for people diagnosed with COVID-19, these attractors shift and become chaotic. This difference in the shape of the attractors allows for classifiers to effectively distinguish between the two classes. The best separation of attractor shapes is observed for the MCHC–MCH pair (Figure 7a) that explains its highest classification ability. For the pairs in Figure 7b–e, shifted cruciform attractors are observed, which also contributes to their effective separation by classifiers. In Figure 7f, two attractors are blurred, but due to their weak intersection, the classification efficiency is high.

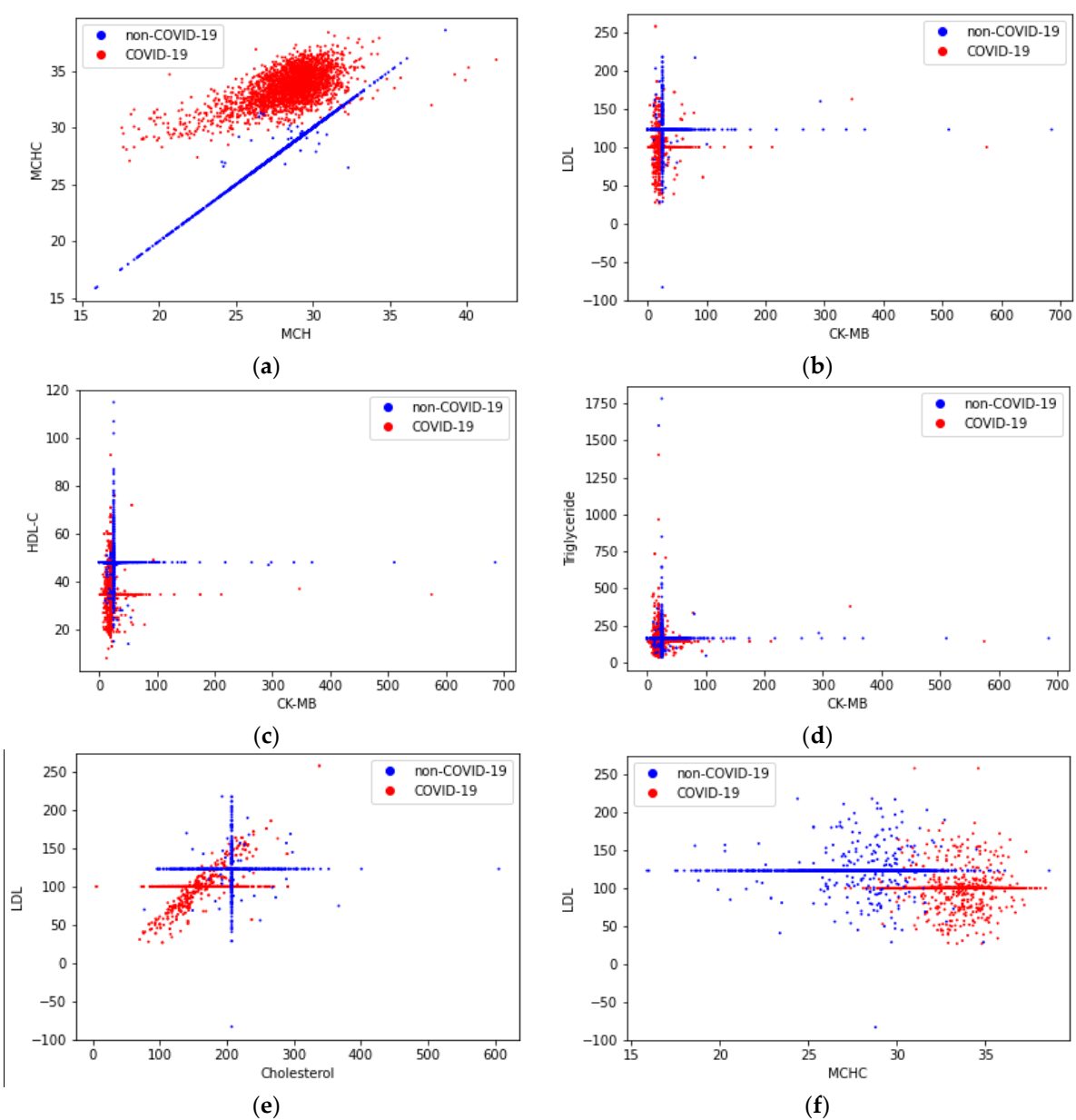

**Figure 7.** Two-dimensional distributions (attractors) of a COVID-19 and non-COVID-19 diagnosis in the coordinates of feature pairs MCHC–MCH (**a**), LDL–CK-MB (**b**), HDL-C–CK-MB (**c**), Triglyceride–CK-MB (**d**), LDL–Cholesterol (**e**), LDL–MCHC (**f**).

When using two features, the maximum accuracy is $A_2$ = 99.81% and this value is lower than when using the 51 features $A_{51}$ = 100%. However, feature reduction is important to simplify the classification of patients in practical terms. More accurate models can be obtained using three features.

*4.2.3. The Study of the Most Significant Combination of three Features of the HGB Model*

Table 8 presents the classification result of the SARS-CoV-2-RBV1 dataset for the HGB model using three input features.

**Table 8.** Classification efficiency of SARS-CoV-2-RBV1 dataset using 3 features for the Histogram-based Gradient Boosting classifier.

| № | First feature | Second feature | Third feature | Average accuracy $A_3$,% |
|---|---|---|---|---|
| 39-48-32 | Cholesterol | Triglyceride | CK-MB | 99.91 |
| 39-36-32 | Cholesterol | HDL-C | CK-MB | 99.91 |
| 43-20-19 | LDL | MCHC | MCH | 99.91 |
| 20-31-19 | MCHC | Amylase | MCH | 99.85 |
| 43-51-32 | LDL | UA | CK-MB | 99.85 |
| 39-20-19 | Cholesterol | MCHC | MCH | 99.83 |
| 48-42-32 | Triglyceride | LDH | CK-MB | 99.83 |
| 36-20-19 | HDL-C | MCHC | MCH | 99.79 |
| 36-42-32 | HDL-C | LDH | CK-MB | 99.79 |
| 43-38-51 | LDL | Cholesterol | UA | 99.79 |
| 20-48-19 | MCHC | Triglyceride | MCH | 99.77 |
| 39-48-31 | Cholesterol | Triglyceride | Amylase | 99.77 |
| 39-36-38 | Cholesterol | HDL-C | Chlorine | 99.75 |
| 36-31-51 | HDL-C | Amylase | UA | 99.75 |
| 39-36-42 | Cholesterol | HDL-C | LDH | 99.75 |
| 20-51-19 | MCHC | UA | MCH | 99.74 |
| 39-48-38 | Cholesterol | Triglyceride | Chlorine | 99.72 |
| 39-31-51 | Cholesterol | Amylase | UA | 99.70 |
| 39-48-42 | Cholesterol | Triglyceride | LDH | 99.66 |
| 48-31-42 | Triglyceride | Amylase | LDH | 99.51 |

An analysis of Table 8 reveals that no new features have been added in the first twenty most accurate models compared to Table 7. The MCH and MCHC features are found only in pairs. With the addition of the third feature, the maximum classification efficiency increased from $A_2$ = 99.81% to $A_3$ = 99.91%.

4.2.. The Study of the Most Significant Combination of 11 Features of the HGB Model

Table 7 and Table 8 include only 11 features: LDL (№43), cholesterol (№39), HDL-C (№36), MCHC (№20), triglyceride (№48), amylase (№31), UA (№51), LDH (№42), CK-MB (№32), ALP (№30) and MCH (№19). The classification accuracy of the HGB model using 11 features was $A_{11}$ = 100%. Therefore, 11 features are sufficient to determine the presence of COVID-19 using machine learning methods based on the histogram-based gradient boosting classifier.

*4.3. LogNNet Implementation on Arduino for Edge Computing*

A compact 77-line LogNNet algorithm was created for diagnosing and predicting COVID-19 disease using routine blood values on an Arduino controller.

LogNNet testing on Arduino revealed an accuracy of $A_{51}$ = 99.7%, which coincides with the accuracy on the model computer program [46]. The classification time for the input vector is about 0.1 s. An estimate of the RAM used by the Arduino controller is shown in Figure 8.

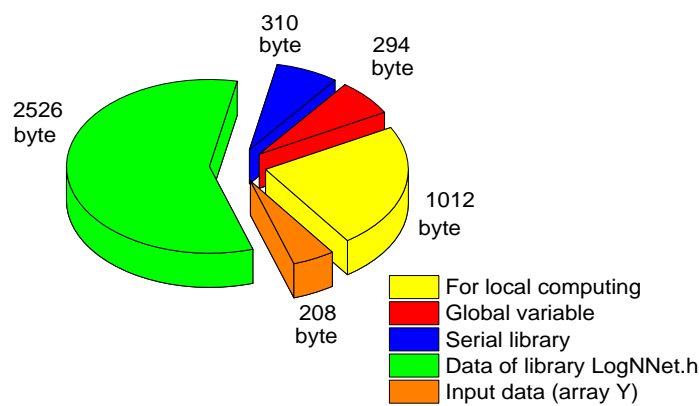

**Figure 8.** Estimation of the RAM used by the Arduino controller when working with the neural network LogNNet 51:50:20:2.

Global variables (arrays $S_h$, $S_{h2}$ and variables) occupy 294 bytes of RAM, and incoming data is written to array $Y$, which occupies 208 bytes. The Arduino uses the Serial system library to operate the serial port. It is loaded at initiation in the "void setup" block and takes 310 bytes of RAM. The data stored in the LogNNet.h library are also loaded into the RAM during the program's initialization and take 2526 bytes, the maximum contribution made by the matrix $W_1$—2142 bytes. For local computations within functions and procedures, at least 1012 bytes must be reserved. The total RAM consumption is of 4350 bytes.

### 4.4. Machine Learning COVID-19 Sensor for IoT

The LogNNet network can be easily imported to various microcontrollers and used to predict a diagnosis based on blood biochemical parameters. However, our experimental results in Sections 3.2 and 3.3 are significantly inferior in accuracy to resource-intensive machine learning algorithms. Therefore, we proposed two architectures of the IoT system (Figure 9), which include an IoT device with LogNNet implementation (edge computing) and a cloud service containing a trained HGB model (AI computing). These configurations implement the prognosis of the disease in offline and online modes with ML sensors for diagnosis of the COVID-19 disease (Sensor 1.0 type).

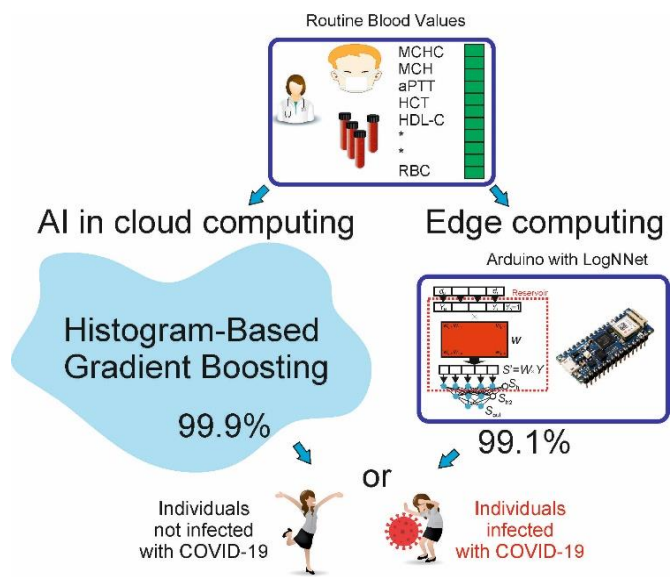

**Figure 9.** Two architectures of the IoT system, which includes an IoT device with LogNNet implementation (edge computing) and a cloud service containing a trained HGB model (AI computing).

In the IoT device, the results of a biochemical blood test are entered manually or transmitted directly from the laboratory equipment. If the cloud service is unavailable or if the blood tests are performed on site using a mobile laboratory in remote areas, the diagnosis is made by the LogNNet network. If the IoT device has access to the network, it sends a network request to the cloud service, wherein the diagnosis is determined using the HGB model. The cloud service sends a response with a diagnosis, which is displayed on the IoT device using an LED indication or on an LCD display.

## 5. Discussion

### 5.1. Analysis of Results from a Medical Perspective

COVID-19 has a higher mortality and infectivity than influenza [3,13]. The disease still causes death and continues to spread [1,6,15]. The use of vaccines did not stop the spread of the disease, and important mutations were detected in the structure of the virus during the epidemic [1]. Most of the infected patients had mild symptoms and a good prognosis. However, some patients developed severe symptoms, such as severe pneumonia, acute respiratory distress syndrome (ARDS) and multiple organ dysfunction syndromes (MODSs) [2,5,24]. A need for studies to determine the prognosis and immune conditions of the COVID-19 disease remains [3,74]. Therefore, the early evaluation of patients who need intensive care and have high mortality expectations as well as the effective identification of relevant biomarkers are important to reduce the mortality of the disease [5,6,25].

Various complications may be encountered during the treatment process of COVID-19 and, therefore, the course of the disease should be predicted earlier [64,77]. It is important to diagnose and predict the prognosis of the disease at an early stage so that the first response to severely infected COVID-19 patients can be conducted properly [2,5].

Although many studies on COVID-19 have been published, the relationships between the pathological aspects of the disease and routine blood values have not been fully determined [77]. Previous studies have reported that changes in many RBVs and hematological abnormalities are observed during the course of the disease [14,77].

In this study, according to the $A_{th}$ threshold classification result based on [12], the most effective features in the diagnosis of the disease were found to be LDL with 96.47%, HDL-C with 94.73%, cholesterol with 94.47% and MCHC with 94.35% (Table 3). Indeed, in previous studies, large changes in these features were reported in severe and fatal COVID-19 patients, and these features may be important biomarkers for the prognosis of the disease [1,2,5,14].

Considering the linear dependency structures of the features among each other, the most effective combinations of dual features in the diagnosis of the disease were obtained and the Pearson correlation values were calculated (Table 4). The highly positive linear correlation structure of some trait pairs with positive and negative individuals was remarkable. The highest positive and negative linearly correlated trait pairs (High–High) were HCT–HGB, MPV–PDW and HCT–RBC (96%, 93% and 87%, respectively). These features vary greatly in severe COVID-19 patients and may be associated with the prognosis and mortality of the disease [1,2,14]. The high positive association of trait pairs expressed as the High–High type with both positive and negative COVID-19 individuals led us to believe that various comorbidities such as hypertension, obesity and diabetes may exist in our negative COVID-19 population. Considering hospital admissions of negative COVID-19 patients, these trait pairs are highly associated not only with COVID-19 but also with various inflammatory syndromes and infections [1,2,78]. Djakpo et al. [79] stated that the abnormalities of HGB, HCT and RBC or anemia observed in patients with comorbidities are due to the inability of the bone marrow to produce enough RBCs to carry oxygen and lung damage caused by COVID-19 which complicates gas exchange.

Considering the relationship of the patients with these features, the presence of possible comorbid conditions prevents erythrocyte production due to existing inflammation. Since the variation in these trait pairs is hypersensitive to the immune response in individuals, these trait pairs were highly correlated with sick and healthy individuals. The MCH–MCHC trait pair was found to be highly positively correlated, especially with healthy individuals, and this pair may be used as an important marker to distinguish healthy individuals in the diagnosis of the disease. Changes in these characteristics may indicate the suppression of lymphocytic and erythrocyte series or platelet and erythrocyte deformities [1]. In addition, in this study, a highly positive association of the MCH–MCV trait pair with COVID-19 was found. Mertoğlu et al. [2], Huyut et al. [14] and Karakike et al. [21] stated that this was due to the decrease in the size of erythrocytes and anisocytosis in patients. The high positive association of the HGB–RBC trait pair with sick individuals may be related to impaired erythropoiesis in the later stages of the disease. The High–High type feature set provides important clues in the isolation of both sick and healthy individuals.

In Table 4, a high (77%) negative relationship between the fibrinogen–LYM feature pair and COVID-19 patients is seen, and we believe that this level of relationship is due to the fibrinogen feature. Indeed, Winata and Kurniawan [28] noted that the degradation product of fibrinogen (FBU) was increased in all patients in the late stage of COVID-19 and that this was significantly associated with coagulation. In addition, the high correlation of the cholesterol–LDL, cholesterol–HDL-C and chlorine-sodium trait pairs (High–Low type in Table 4) with sick individuals showed that these trait pairs were important markers in identifying sick individuals. Fang et al. [80] and Mertoğlu et al. [1] stated that this feature set may be associated with multi-organ involvement in COVID-19 and the widespread distribution of angiotensin-converting enzyme receptors in the body.

The fact that the Low–High trait pair MCHC–MCV was found to be highly positively correlated with COVID-19 negative individuals in Table 4 suggested the importance of the size of erythrocyte and anisocytosis in healthy individuals [80]. In addition, the functional properties of the ALT–AST, eGFR–Urea and D-Bil–T-Bil pairs were found to be important markers in the isolation of COVID-19 negative individuals. Mertoğlu et al. [1], Huyut et al. [14] and Zhou et al. [27] stated that the decrease in ALT, AST, GGT, total bilirubin and eGFR indicated that the patients had serious damage to organs such as pancreas and kidney. In another study, Bertolini et al. [81] stated that AST, GGT, ALP and bilirubin may be frequently elevated in COVID-19 and that the main underlying causes of this condition may be hyper inflammation and thrombotic microangiopathy. In addition, the high positive correlation of the INR–PT trait pair with negative COVID-19 individuals suggested that it is important to monitor these individuals for the development of disseminated intravascular coagulopathy and acute respiratory distress [80,82].

In this study, 13 popular classifier machine learning models and the LogNNet neural network model were run on 51 routine blood values to detect patients infected with COVID-19. Histogram-based gradient boosting (HGB) was the model with the fastest and highest accuracy in determining the diagnosis of the disease (accuracy: 100%, time: 6.39 sec).

For the HGB model using a single input feature ($A_1$), the most effective features in the diagnosis of the disease were LDL 96.87% (№43), cholesterol 9507% (№39), HDL-C 94.99% (№36), MCHC 94.35% (№20), triglyceride (№48) and amylase (№31) (Table 6). For the HGB model using the dual entry feature ($A_2$), the most effective trait pair found in the diagnosis of the disease was MCHC–MCH ($A_2$ = 99.81) (Table 7). The success of MCH as a single-entry feature in the diagnosis of the disease is low ($A_1$ = 52.13). Huyut and Velichko [12] found an accuracy rate of 99.1% in the diagnosis of the disease by running the MCHC–MCH features with LogNNet. The HGB model operated with MCHC–MCH was found to be more successful than the LogNNet model in the diagnosis of the disease.

Since low values of MCH and high values of MCHC were associated with COVID-19 [83], it was expected that the use of these two features together in the diagnosis of the disease would produce higher classification success. The most effective dual trait pairs (Table 7) were similar to the most effective single traits (Table 6) for the HGB model for the diagnosis of the disease. This provides important information about the functional properties of the binary trait pairs obtained with the HGB model in the diagnosis of the disease. Six basic features, that is LDL (№43), cholesterol (№39), HDL-C (№36), MCHC (№20), triglyceride (№48) and amylase (№31), among the combinations of binary features used in the diagnosis of the disease and four features, that is UA (№51), LDH (№42), CK-MB (№32) and ALP (№30), that most frequently pair with these features are given in Figure 6. The main feature LDL (№43) generated the largest number of effective pairs for classification. The effectiveness of these feature pairs (Table 7) in detecting patients is visualized in two-dimensional space (Figure 7). Classification is most clearly visible in the MCHC–MCH pair (Figure 7a), which explains the higher classification ability.

In the binary feature combinations used by HGB in the diagnosis of the disease, the maximum accuracy was $A_2$ = 99.81% which is slightly lower than the use of 51 features ($A_{51}$ = 100%). However, feature reduction provides more cost effective and rapid results in interpreting the classification of patients from a practical point of view and identifying the most effective features. The highest classification success obtained for the HGB model using three feature combinations was $A_3$ = 99.91 (Table 8).

Analysis of Table 8 showed that no new features were added to the top twenty models with the highest accuracy compared to Table 7. The binary combinations in Table 7 were sufficient for the diagnosis of the disease. In addition, the co-existence of MCH and MCHC features in all combinations reveals hidden association structures between these features and contains important clues in the diagnosis of the disease.

In this study, the most important 11 biomarkers were found with the HGB model used to determine the diagnosis of the disease, and with these features, all patients and healthy individuals were correctly identified with high performance (A11 = 100%). In addition, the importance of various combinations of these features in the diagnosis of the disease was recognized. The performance of these 11 features, namely LDL (№43), cholesterol (№39), HDL-C (№36), MCHC (№20), triglyceride (№48), amylase (№31), UA (№51), LDH (№42), CK-MB (№32), ALP (№30) and MCH (№19) and their various combinations in the diagnosis of the disease was higher than the individual performances, suggesting that there is a high level of confidential information between these feature combinations and COVID-19.

Kocar et al. [84] and Zinellu et al. [85] presented evidence of significant changes in the lipid profile of severe COVID-19 patients, particularly in total cholesterol, LDL and HDL-C concentrations. They also reported that increased cholesterol concentrations in the cell membrane increased the binding activity of SARS-CoV-2, facilitated membrane fusion and enabled the successful entry of the virus into the host. Therefore, Kocar et al. [84] and Wei et al. [86] indicated that total cholesterol, LDL and HDL-C characteristics may aid in early risk stratification and clinical decisions. However, conflicting results have been reported for changes in triglyceride levels of severe COVID-19 patients [85]. Stephens et al. [87] stated that in severe COVID-19 patients, the elevated serum amylase value is often not attributable to acute pancreatitis or a clinically significant pancreatic injury, but is more likely to be a nonspecific manifestation of shock/critical illness. Mao et al. [83] stated that changes in leukocytes, neutrophils, lymphocytes, platelets, hemoglobin levels, MCV and MCHC are generally associated with lung involvement, oxygen demand and disease activity. They also noted that high MCV and low MCHC are associated with advanced anemia and are independent predictors of disease worsening [83].

Wu et al. [88] stated that an increase or decrease in LDH is indicative of radiographic progression or improvement. They also demonstrated the potential usefulness of serum LDH as a marker for assessing clinical severity, monitoring treatment response and thus aiding risk stratification and early intervention in COVID-19 pneumonia. Hu et al. [89] stated that SARS-CoV-2 infection is associated with low serum uric acid (SUA) levels, and this feature may be an independent risk factor for the disease. They also noted that male patients with COVID-19 accompanied by low SUA levels are at higher risk of developing severe symptoms than those with high SUA levels at admission. Zinellu et al. [90] found that high CK-MB concentrations were significantly associated with severe morbidity and mortality in COVID-19 patients. They stated that this biomarker of myocardial damage may be useful for the classification of patients with severe COVID-19, and that high CK-MB values may reflect excessive inflammation status. They also stated that the evaluation of CK-MB in COVID-19 patients provides specific clinical information for early risk stratification, independent of myocardial necrosis and cardiac complications. Afra et al. [91] showed the incidence of abnormal liver tests in severe COVID-19 patients and reported the association of elevated AST, ALT and total bilirubin levels with liver injury in severe COVID-19 patients [13,91,92]. However, conflicting results have been reported regarding the ALP levels of mild and severe COVID-19 patients [91]. In addition, Afra et al. [91] showed that elevated liver enzymes can effectively predict hospital-critical COVID-19 cases.

The accuracy of ML algorithms is difficult to determine when used without any physician input [93,94]. A major limitation of ML is that it is difficult to explain how these algorithms arrive at their conclusions [95]. An ML algorithm can be likened to a black box that takes inputs and produces outputs without any explanation as to how it produces the outputs [94].

Additionally, if an algorithm misdiagnoses a malignant lesion, the algorithm cannot explain why it chose a particular diagnosis [94,95]. While the printouts can aid interpretation, it can be a potential danger and problem to the patient if the model fails to explain to a patient why he or she has diagnosed a lesion as benign or malignant, or how it has chosen a particular treatment [94].

Physician interpretation is necessary for choosing a diagnosis or treatment. In addition to the black box nature of these algorithms, machine learning is also prone to the "garbage in, garbage out" motto [94]. This maxim indicates that the quality of the dataset input determines the quality of the output. Therefore, if the data inputs are badly labeled, the outputs of the algorithm will reflect these inaccuracies [93–95].

In addition, all the devices should be evaluated in prospective clinical trials and made publicly available in the peer-reviewed literature.

### 5.2. Analysis of Results from IoT Perspective

The aim of this study is the feasibility analysis of a fast, reliable and cost-effective digital tool for the diagnosis of COVID-19 based on the RBV values measured at admission. The proposed solution is based on the concept of ML sensors for diagnosis of the COVID-19 disease (Sensor 1.0 type). The concept makes a step towards "smart sensorics of human" with promising opportunities for AI applications in healthcare.

In our study, we are not targeting IoT systems for telemedicine, wherein any procedure is performed by a physician using telecommunication means of transmitting medical data. Solutions of clinical telemedicine are subject to strict certification. Telemedicine is prescribed by a doctor and is administered via a medical device (never via a smartphone). Instead, we focus on promising IoT systems for telehealth/telecare and mobile health (m-Health) [45]. First, data from ML sensors support the prognosis of the disease in offline and online modes. Second, ML sensors can be used in AAL and other IoT environments to support a person in his/her everyday life. Importantly, COVID-19 is not the only disease to apply ML sensors in IoT systems.

AI methods become effective for the prognosis of various diseases. COVID-19 has opened the new era for AI methods to mitigate future pandemics. The rapidly growing number of publications confirms the potential of ML sensors for collecting datasets for further analysis with AI methods. Predictive analytics uses available retrospective data and various predictive models (including ML-constructed) to aid in answering the question "What could happen?". Prognosis from the sensed data is required not only for clin-ical medicine (to support clinical medical decisions). Managerial predictive analytics supports managers in healthcare at various levels to assess possible scenarios for the development of diseases, the budgets of medical organizations, the need for medicines, etc.

In AAL, ML sensors are useful in personal use as digital assistance (recommendations, including prognosis). In fact, the five natural human sensors (vision, hearing, touch, smell, taste) are extended by ML sensors. A person can develop health insight from their own RBVs in real time or collect the data for retrospective analysis. Humans themselves can act as complicated sensors [44]. A human traditionally finds a way to enhance her/his function, e.g., glasses (optical tool) to advance the vision or thermometer (physical tool) to regularly sense body temperature. Now the era of digital tools for personal health assistance is coming.

The implemented prototype of the diagnosis tool demonstrates that the LogNNet network can be imported to various microcontrollers. Many IoT devices can be made smarter, opening a way to develop advanced AmI healthcare with essential parts of IoT and edge intelligence [96]. A LogNNet-equipped ML sensor can be effectively employed in future IoT applications for healthcare and for other problem domains that require active digitalization and emerging AmI methods [46].

The LogNNet network can be used to predict a diagnosis based on blood biochemical parameters. This result is an important step in smart human sensors for IoT application, as the COVID-19 status and other blood-related health parameters are difficult to analyze on the IoT edges (in contrast to more widespread parameters, such as temperature or heartbeat) [97–99]. Our approach is applicable to the development of personalized bionic systems (smart suit for a person or AmI environment with people), wherein disease status recognition is a regular digital service for healthcare or well-being applications in everyday life [45].

Although the small IoT devices cannot provide such high accuracy as resource-intensive ML algorithms on powerful computer systems, AAL systems are intended for everyday life settings (e.g., at home, workspace, outdoor). Where strict medical decisions and critical medical support are not mandatory, the digital services may provide attention points and optional recommendations for personal use. We believe this type of smart human sensors will soon diffuse from the restricted medical lab setting toward the wide market of smart consumer electronics and digital services [100].

## 6. Limitations of the Study

The data primarily represent a single institution (EBYU-MG) and the Turkish population. Secondly, our dataset does not include comorbidities of patients and other diagnostic information of patient subgroups. In practice, the data in retrospective studies collected in a certain period cannot meet all data sample requirements. We suggest the findings in this study be supported by a retrospective cohort study setup.

## 7. Conclusions and Future Studies

Determining a COVID-19 infected status with diagnostic tests and imaging results is costly and time-consuming. If this process is prolonged, the patient's health may be at greater risk by being exposed to different complications. This study provides a fast, reliable and cost-effective alternative mobile tool for the diagnosis of COVID-19 based on the RBVs measured at the time of admission.

In this study, 13 popular classifier machine learning models and the LogNNet neural network model were run on 51 routine blood values to detect patients infected with COVID-19. The histogram-based gradient boosting (HGB) model was the most successful classification model in terms of accuracy and time in detecting the diagnosis of the disease (accuracy: 100%, time: 6.39 s). In addition, the absence of any normalization method and additional feature selection procedure for the HGB model contributes to the speed and efficiency of the model.

The eleven most important biomarkers in the diagnosis of the disease were found with the HGB classifier: LDL (№43), cholesterol (№39), HDL-C (№36), MCHC (№20), triglyceride (№48), amylase (№31), UA (№51), LDH (№42), CK-MB (№32), ALP (№30) and MCH (№19). Using only these 11 RBVs features, the HGB model accurately detected all COVID-19 patients ($A_{11}$ = 100%).

The high accuracy of the single, double and triple combinations of these 11 features selected by the HGB model in the diagnosis of the disease showed the importance of these features in the diagnosis of the disease. In addition, the performance of double and triple combinations of these features in the detection of sick and healthy individuals was higher than the individual performances, suggesting that there is a high level of hidden information between these blood feature combinations and COVID-19.

The HGB model reveals that 11 features are sufficient for the diagnosis of the presence of COVID-19 using the HGB classifier. These features and their binary combinations are an important source of variation in the diagnosis of COVID-19. We propose to use these features and their binary combinations to be run with HGB as important biomarkers in the diagnosis of the disease.

The study results can be effectively used in IoT medical edge devices with low RAM resources, ML sensors, portable point-of-care blood testing devices [101], decision support systems, telecare and m-Health. This opportunity empowers the development of many innovative applications for predictive analytics in clinical MIS or everyday AAL systems.

The artificial intelligence models for the early prediction of the diagnosis and progression of COVID-19 and other diseases produce satisfactory results. Future artificial intelligence studies for the early diagnosis and prognosis of fatal, costly and severe diseases will ease the burden of healthcare professionals and increase patient comfort. In addition, the use the physiological, comorbidity and demographic features of the patients together with the RBVs data may provide interesting insights. Testing the results of this study on multi-racial, multi-center and larger patient groups may improve the generalizability of the findings. In this context, this study may pave the way for many exciting subsequent investigations.

**Author Contributions:** Conceptualization, M.T.H. and A.V.; methodology, A.V., M.T.H., B. M., Y.I. and D.K.; software, A.V., M.T.H., B.M. and Y.I.; validation, M.T.H. and A.V.; formal analysis, A.V., M.T.H., B.M., Y.I. and D.K.; investigation, A.V.; resources, M.T.H.; data curation, M.T.H.; writing—original draft preparation, A.V., M.T.H., B.M., Y.I. and D.K.; writing—review and editing, A.V., M.T.H., B.M., Y.I. and D.K.; visualization, A.V. and B. M.; supervision, D.K.; project administration, D.K.; funding acquisition, D.K. All authors have read and agreed to the published version of the manuscript.

**Funding:** The research is implemented with financial support by Russian Science Foundation, project no. 22-11-20040 (https://rscf.ru/en/project/22-11-20040/ (accessed on 14.10.2022)) jointly with Republic of Karelia and funding from Venture Investment Fund of Republic of Karelia (VIF RK).

**Institutional Review Board Statement:** The dataset used in this study was collected in order to be used in various studies in the estimation of the diagnosis, prognosis and mortality of COVID-19. The necessary permissions for the collected dataset were given by the Ministry of Health of the Republic of Turkey and the Ethics Committee of Erzincan Binali Yıldırım University. This study was conducted in accordance with the 1989 Declaration of Helsinki. Erzincan Binali Yıldırım University Human Research Health and Sports Sciences Ethics Committee Decision Number: 2021/02-07.

**Informed Consent Statement:** In this study, a dataset including only routine blood values, RT-PCR results (positive or negative) and treatment units of the patients was downloaded retrospectively from the information system of our hospital in a digital environment. A new sample was not taken from the patients. There is no information in the dataset that includes identifying characteristics of individuals. It was stated that routine blood values would only be used in academic studies, and written consent was obtained from the institutions for this. In addition, therefore, written informed consent was not administered for every patient.

**Data Availability Statement:** The data used in this study can be shared with the parties, provided that the article is cited.

**Acknowledgments:** We thank the method of Erzincan Mengücek Gazi Training and Research Hospital for their support in reaching the material used in this study. Special thanks to the editors of the journal and to the anonymous reviewers for their constructive criticism and improvement suggestions.

**Conflicts of Interest:** The authors declare no conflict of interest.

## Appendix A

**Algorithm A1.** LogNNet neural network executable code on Arduino Nano IoT 33.

```
1   #include "LogNNet.h"
2
3   float Y[S+1];
4   float Sh[P+1];
5   float Sh2[M+1];
6
7   int i = 0;
8   String data;
9
10  float Fun_activ(float x) {
11     return 1 / (1 + exp(-1*x));
12  }
13
14  void Reservoir(float *Y) {
15     long W = C;
16     Sh[0] = 1;
17     for (int j = 1; j <= P; j++) {
18        Sh[j] = 0;
19        for (int i = 0; i <= S; i++) {
20           W = (D - K * W) % L;
21           Sh[j] = Sh[j] + ((float)W/L) * Y[i];
22        }
23        Sh[j] = ((Sh[j] - (float)minS[j-1]/
24           scale_factor) / ((float)(maxS[j-1]
25           - minS[j-1])/scale_factor)) - 0.5
26           - (float)meanS[j-1]/(scale_factor*10);
27     }
28  }
29
30  void Hidden_Layer() {
31     Sh2[0] = 1;
32     for (int j = 1; j <= M; j++) {

40
41  byte Output_Layer() {
42     float Sout[N+1]; byte digit = 0;
43     for (int j = 0; j <= N; j++) {
44        Sout[j] = 0;
45        for (int i = 0; i <= M; i++)
46           Sout[j] = Sout[j] + Sh2[i] *
47              ((float)W2[i][j]/scale_factor);
48        Sout[j] = Fun_activ(Sout[j]);
49     }
50     for (int j = 0; j <= N; j++) {
51        if (Sout[j] > Sout[digit])
52           digit = j;
53     }
54     return digit;
55  }
56
57  void setup() {
58     Serial.begin(115200);
59  }
60
61  void loop() {
62     if (Serial.available() > 0) {
63        data = Serial.readStringUntil('T');
64
65        if (data != "FN") {
66           Y[i] = data.toFloat();
67           i++;
68        }
69        else {
70           i = 0;
71           Reservoir(Y);
```

```
33   Sh2[j] = 0;                                72      Hidden_Layer();
34   for (int i = 0; i <= P; i++)               73      byte Digit = Output_Layer();
35     Sh2[j] = Sh2[j] + Sh[i] *                74      Serial.print(String(Digit));
36         ((float)W1[i][j]/scale_factor);      75    }
37   Sh2[j] = Fun_activ(Sh2[j]);                76  }
38   }                                          77 }
39 }                                            78
```